%% file: main.tex
\documentclass[nonacm,10pt, sigplan]{acmart}

\AtBeginDocument{%
  }

\setcopyright{acmlicensed}
\copyrightyear{2018}
\acmYear{2018}
\acmDOI{XXXXXXX.XXXXXXX}
\acmConference[Conference acronym 'XX]{Make sure to enter the correct
  conference title from your rights confirmation email}{June 03--05,
  2018}{Woodstock, NY}
\acmISBN{978-1-4503-XXXX-X/2018/06}

\usepackage{tikz}
\usepackage{algorithm, algpseudocode}
\usepackage{algpseudocode}
\usepackage{hyperref}
\usepackage{mdframed}
\usepackage{tcolorbox}
\usepackage{overpic}
\usepackage{amsmath}
\usepackage{pifont}
\usepackage{multirow}
\usepackage{enumitem}
\usepackage{subfigure}
\usepackage{makecell}
\usepackage{xspace}
\newcommand{\Tab}{\hspace{0.5cm}}
\newcommand{\beyond}{\textsc{HGCA}\xspace} 
\newcommand{\blacknumber}[1]{
    \tikz[baseline=(char.base)]{
        \node[shape=circle,fill=black,inner sep=0.5pt] (char) {\footnotesize \textcolor{white}{\bfseries #1}};
    }
}
\begin{document}
\pagestyle{plain}
\title{\beyond: Hybrid GPU-CPU Attention for Long Context LLM Inference}

\author{
 {\rm Weishu Deng, Yujie Yang, Peiran Du, Lingfeng Xiang,\\
 Zhen Lin, Chen Zhong, Song Jiang, Hui Lu, Jia Rao } \\
 The University of Texas at Arlington 
 }

\renewcommand{\shortauthors}{Trovato et al.}

\begin{abstract}

Scaling inference for large language models (LLMs) is increasingly constrained by limited GPU memory, especially due to growing key-value (KV) caches required for long-context generation. While existing approaches offload KV caches to CPU memory or apply sparse attention to reduce GPU load, they often underutilize CPU compute resources and compromise accuracy. We present \beyond, a hybrid CPU-GPU attention mechanism that enables scalable, high-throughput LLM inference with near-full attention quality. \beyond performs dense attention on recently generated KV entries retained in GPU memory and parallel sparse attention on selected, salient KV entries in CPU memory. The attention outputs are efficiently merged using log-sum-exp fusion, minimizing PCIe transfer overhead. \beyond also introduces a fine-grained, per-head sparsification strategy optimized for CPU execution, preserving contextual relevance while reducing computation. Our implementation seamlessly integrates into existing LLM frameworks without requiring model retraining. Experiments across diverse models and workloads show that \beyond achieves superior scalability, supports longer sequences and larger batch sizes, and outperforms existing sparse attention baselines in both performance and accuracy -- all on commodity GPU hardware.

\end{abstract}

\maketitle

\section{Introduction}
Transformer-based large language models (LLMs) have shown remarkable capabilities in a wide range of applications, including chatbots~\cite{dam2024complete}, code assistance~\cite{roziere2023code}, text generation and summarization~\cite{pilault2020extractive}. Recent studies \cite{feng2024towards, wei2022chain} show that providing step-by-step Chain-of-Thought (CoT) enhances LLM’s ability to solve complex tasks involving reasoning and mathematical problem-solving. Such inference-time scaling techniques, including CoT prompting, sampling with majority voting~\cite{leviathan2023fast}, and tree-search-based sequence generation~\cite{zheng2023efficiently}, demand substantial transient states, typically in the form of key-value (KV) caches, during LLM inference. The size of the KV cache scales with input and output sequence lengths, as well as batch size, and may exceed the size of the model weights, becoming the primary contributor to memory consumption in LLM serving. As researchers strive to make LLMs accessible to small institutions and individual users, efficiently serving LLM inferences on single or multiple commodity GPUs has become increasingly important. 

While many pre-trained, open-source LLMs can accommodate their model weights within the memory capacity of commodity GPUs, such as the NVIDIA A6000, the KV cache can grow without bound, eventually depleting available GPU memory. To enable long-sequence generation and support concurrent inference requests, modern LLM serving systems~\cite{kwon2023efficient} incorporate mechanisms for offloading KV caches to CPU memory.
Unlike model weights, which must be fully loaded and used to transform input representations throughout inference, the intermediate results from an LLM's attention layers (keys and values) can be stored for reuse in the KV cache or recomputed for each newly generated token. The key challenge in KV cache management is efficiently reusing stored key-value entries when the cache exceeds the available GPU memory capacity. 

Although a significant portion of the KV cache can be offloaded to CPU memory, it is generally believed that attention computation on the KV cache should be performed on the GPU to fully utilize its computational power. Full attention-based KV management~\cite{sheng2023flexgen} requires the entire KV cache stored in CPU memory to be transferred to the GPU for attention computation. However, loading the KV cache over the PCIe bus -- limited to up to 32 GB/s on commodity GPUs with 16 PCIe 4.0 lanes -- introduces a performance bottleneck and disrupts attention computation on the GPU. In contrast, sparse attention~\cite{zhang2024h2o} attends a subset of KV cache entries with the highest attention scores and only keeps these entries in GPU memory. While sparse attention reduces the need to frequently load KV cache from CPU memory, it depends on accurately identifying important KV entries. Excluding critical KV entries due to a reduced attention scope can degrade the accuracy of LLM inference. As LLMs scale and content generation becomes more complex, the demand for intermediate data storage increasingly outpaces available GPU memory, resulting in reduced inference throughput and potential accuracy degradation.

In this paper, we seek to develop an attention mechanism for LLM inference that simultaneously achieves good throughput, high accuracy, and (almost) unlimited~\footnote{The scale of LLM inference due to KV caching is not limited by the capacity of GPU memory.} scalability to long sequence outputs and large batch sizes. This paper seeks to answer a fundamental question -- {\em Given an LLM model, what is the highest performance that can be achieved on a single consumer-class GPU without undermining model accuracy or requiring algorithmic changes to the attention mechanism?} Our work is motivated by two important observations. {\em First}, unlike LLM training, LLM inference often exhibits low operational density~\cite{ye2024accelerating} -- defined as the number of operations per byte of memory traffic -- offering opportunities for offloading a portion of attention computation to the CPU without incurring drastic performance degradations. The performance discrepancies between the computational power (i.e., TFLOPS) of CPUs and consumer-class GPUs are still significant, while the gap in memory bandwidth is much narrower. For example, the gap in TFLOPS for FP16 in the Intel Xeon Gold 6430 processor and NVIDIA A6000, a widely available and adopted GPU for LLM serving, is at least an order of magnitude (1.229 TFLOPS vs. 38.7 TFLOPS). In contrast, the CPUs can achieve up to 500 GB/s memory bandwidth with all 32 DDR slots fully populated, whereas the NVIDIA RTX A6000 GPU's GDDR6 memory can only deliver up to 768 GB/s. {\em Second}, CPUs have advanced control logic that efficiently handles complex sparsification algorithms with intricate control flow, helping close the performance gap in attention computation.

\begin{figure}[!ptb]
\setlength{\abovecaptionskip}{0.1cm}
\setlength{\belowcaptionskip}{-0.01cm}           
\centering
\includegraphics[width=0.7 \columnwidth]{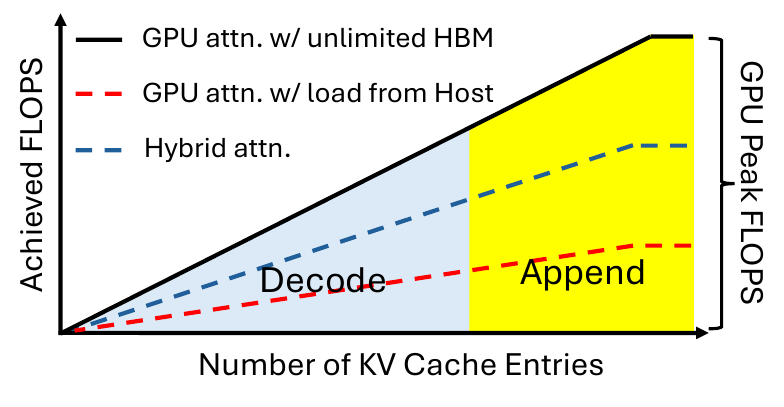}
\caption{Roofline model of attention stages in LLM serving.}
\label{figs:roofline-model}
\end{figure}
 

Figure~\ref{figs:roofline-model} shows a roofline model of attention operators in LLM inference~\cite{ye2024accelerating}. The {\em decode} and majority {\em append} stages are memory-bound, where the CPU can potentially keep up with the GPU during attention computation. Ideally, optimal attention performance is achieved via {\em GPU-only attention} with unlimited GPU memory, always underneath the peak GPU bandwidth ceiling. {\em GPU attention with CPU offloading} (represented by the red dotted line) allows the KV cache size to exceed GPU memory capacity but suffers from performance degradations due to the transfer of the KV cache between the CPU and GPU over PCIe, leaving the GPU idle during the transfer. Alternatively, this work explores {\em hybrid attention} to leverage the combined bandwidth of CPU and GPU memory, achieving the aggregated computational power of both devices.

We propose \beyond, a hybrid CPU-GPU attention mechanism that partitions computation across devices with distinct runtime key-value management. To streamline attention computation on the two heterogeneous devices, \beyond performs full and sparse attention on the GPU and CPU, respectively. Full attention operates on KV entries stored in GPU memory, corresponding to recently generated tokens, whereas sparse attention selectively focuses on high-scoring KV entries residing in CPU memory. The attention output on the CPU is then transferred to the GPU, where the two partial attention outputs are aggregated using log-sum-exp fusion. Furthermore, \beyond leverages multiple CPU threads to perform per-head sparse attention over only the most critical KV entries, effectively preserving contextual information during inference and maintaining high accuracy. 

\beyond's hybrid attention offers two advantages. {\em First}, CPU attention not only contributes computationally to the overall attention computation but also eliminates the cross-device data transfer bottleneck -- the partial attention results are orders of magnitude smaller than raw KV caches and can be efficiently transferred to GPU memory via zero-copy transfer. {\em Second}, \beyond's head-granular sparse attention more accurately identifies critical KV entries compared to the widely used Top-K sparse attention. Experimental results with three representative LLMs show that 1) \beyond scales to much larger LLMs, longer sequence outputs, larger batch sizes on a single or multiple commodity GPUs; 2) \beyond achieves accuracy close-to or better than full attention while significantly outperforming existing sparse attention methods; 3) \beyond is implemented as a drop-in replacement for existing attention layers and can be seamlessly integrated into any LLM with minimal engineering effort.

\begin{figure*}[ptb!]
\centering
\includegraphics[width=0.95\textwidth]{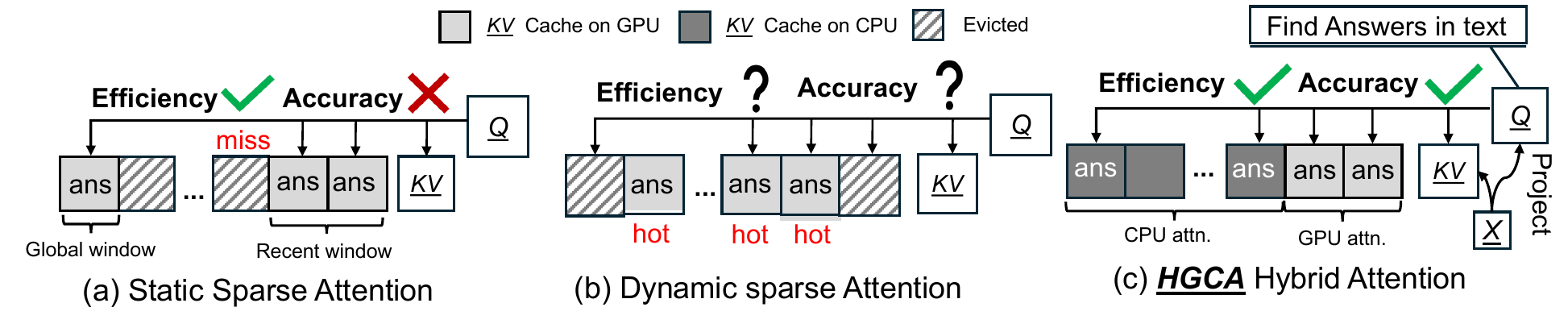}
\caption {The comparison of different attention mechanisms.}
\label{figs:comparison}
 
\end{figure*}

\section{Background and Motivation}
\label{sec:motive}

The attention mechanism is central to LLMs, enabling them to capture complex semantic relationships by referencing previously generated intermediate results. During autoregressive generation, input embeddings from the prompt pass through multiple layers, with KV states cached at each layer for reuse in future token generation.

LLM generation can be formally categorized into three stages—prefill, append, and decode—based on the ratio of query tokens to KV cache entries. In the prefill stage, each input token attends exclusively to newly generated KV states, resulting in a 1:1 query-to-KV ratio. This configuration maximizes GPU utilization, rendering the stage compute-bound. In contrast, the append and decode stages are characterized by substantially lower query-to-KV ratios, leading to reduced KV reuse and increased memory access, which shifts the bottleneck to memory bandwidth. These stages are prevalent in practical scenarios such as prompt extension and autoregressive token generation, particularly in chain-of-thought reasoning, where iterative refinement of outputs is critical for achieving higher accuracy. In the following sections, we provide a detailed explanation of KV cache construction and analyze the performance characteristics of both full and sparse attention mechanisms, with a primary focus on decode and append operations.


\subsection{Standard Full Attention} 

Standard full attention ensures that all KV entries participate in the attention process, preventing any important information from being overlooked. The following equations outline the decode and append processes for LLM inference, as described in~\cite{ye2024accelerating}.
During inference, the incoming hidden states are projected into query, key, and value vectors, which are then transposed to the shape $\mathbb{R}^{h \times N \times d}$, where $h$, $N$, and $d$ represent the number of attention heads, the length of the incoming sequence, and the hidden dimension, respectively.
The incoming $KV$s are subsequently concatenated with the previous $K_{cache}$ and $V_{cache}$, resulting in an expanded $K'V'$ size $N' = N+N_{cache}$,  as shown below. 
\[
K = concat(K_{cache},K)    \quad 
V = concat(V_{cache},V) \in \mathbb{R}^{h \times N' \times d} 
\label{eq:kv}
\] 

Each query vector assesses the relevance of corresponding key-value pairs by computing normalized attention scores using the softmax function applied independently across its head dimension, where higher scores indicate greater influence on the output. These attention scores are subsequently used to compute a weighted sum of the value vectors for each query, producing an output matrix $O$ with the same shape as the query matrix $Q$.

\[
S = QK^T     \quad 
A = softmax(S)  \in \mathbb{R}^{h \times N \times N'} , \quad
O = AV \in \mathbb{R}^{h \times N \times d}
\] 

Queries are discarded after processing, while keys and values can be cached for future reuse. During the decode/append stages, only $N$ output tokens are generated, yet it is necessary to repeatedly load $N’$ KV pairs for the softmax and accumulation operations. Typically, $N \ll N’$, causing memory bandwidth to become a bottleneck, while compute resources remain underutilized. As $N’$ grows beyond the GPU’s memory capacity, offloading KV pairs to host memory becomes a natural solution. However, the significantly lower PCIe bandwidth, several orders of magnitude below that of on-device memory, hampers attention efficiency.




 
\subsection{Sparse Attention}
During attention, many KV pairs receive negligible softmax weights and can be omitted with minimal impact on accuracy. Efficiently identifying and focusing on the most salient KV entries reduces computation, memory usage, and data transfer overhead. As shown in Figure~\ref{figs:comparison}, existing approaches fall into two categories: static and dynamic sparse attention.

{\noindent \bf Static sparse attention} selects important KV entries based on empirically observed patterns and retains only these in GPU memory. The relevance of a KV entry often correlates with its absolute and relative position. Longformer~\cite{beltagy2020longformer} highlights the importance of recent entries within a fixed window, while StreamLLM~\cite{xiao2023efficient} identifies early-sequence tokens (``attention sinks'') as critical for generating accurate outputs. As illustrated in Figure~\ref{figs:comparison}(a), static approaches typically combine a recent window and a global window to select relevant KV entries without runtime overhead.

{\noindent \bf Dynamic sparse attention} identifies salient KV entries at runtime, as shown in Figure~\ref{figs:comparison}(b), enabling context-aware adaptation and reducing the likelihood of excluding important tokens. H2O~\cite{zhang2024h2o} employs a least-recently-used (LRU) policy to track recently accessed entries, assuming temporal locality in attention patterns. Quest~\cite{tang2024quest} performs block-wise scanning to balance accuracy and latency, while Infinigen~\cite{lee2024infinigen} leverages previous queries to asynchronously predict future KV accesses. However, runtime identification introduces computational overhead, which may offset the performance gains of sparsification.

Most sparse attention schemes fix the number of selected KV entries (top-$k$). Twilight~\cite{lin2025twilight} improves upon this by dynamically selecting top-$P$ entries per layer, adapting to context length. However, fine-grained KV tracking per head is impractical on GPUs due to control-flow inefficiencies and uncoalesced memory access, especially in large models.

\begin{figure}[t]
\setlength{\abovecaptionskip}{0.1cm}
\centering
\includegraphics[width=0.9 \columnwidth]{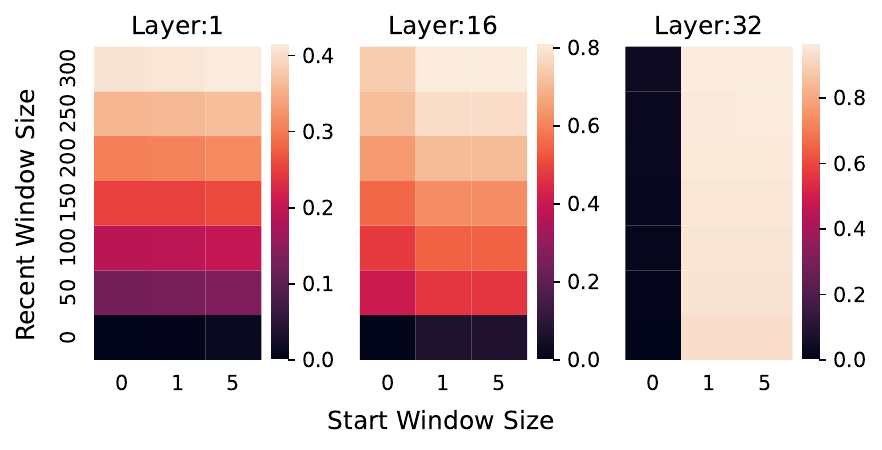}
\caption{The heatmap of cumulative attention weights, aggregated over various starting and recent window, where values closer to 1 indicate high influence on attention output.}
\label{figs:start-recent-weight-sum}
 
\end{figure}
\subsection{Attention Analysis} 
Next, we provide a detailed analysis of attention computation using the OPT-6.7B model on the WikiText dataset. Similar attention patterns were consistently observed across other open-source LLMs and datasets. We focus our analysis on the patterns of KV attention scores across different layers and heads, the spatial and contextual locality between KVs, and the performance implications of KV cache offloading.

Figure \ref{figs:start-recent-weight-sum} draws the heatmap of cumulative attention scores for entry, middle, and exit layers. The cumulative attention scores were aggregated over a recent window and a start window, which measures the effectiveness of sparse attention with a limited scope. As shown in Figure~\ref{figs:start-recent-weight-sum}, attention scores are uniformly distributed in entry layers but become increasingly skewed towards the exit layer. For example, in the OPT-6.7B model, the first token in the input prompt becomes the dominating token in layer 32. This observation suggests that sparse attention should be dynamically adjusted for each layer. Additionally, static sparse attention and the baseline dynamic sparse attention based on top-$k$ KVs are likely to miss important KV entries.

\begin{figure}[t]
\setlength{\abovecaptionskip}{0.1cm}
\setlength{\belowcaptionskip}{-0.05cm}             
\centering
\includegraphics[width=0.9 \columnwidth]{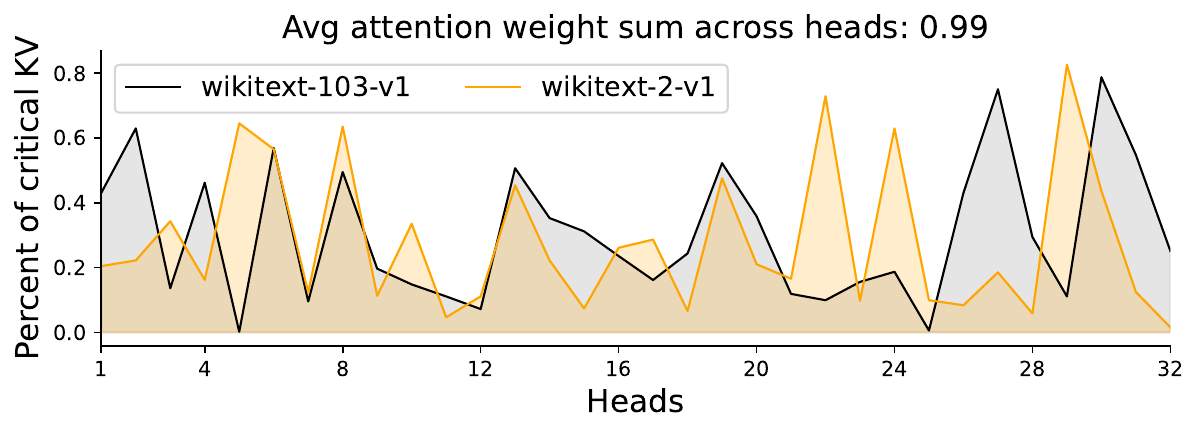}
\caption{The percentage of KV entries required to achieve 99\% of the cumulative attention score per head in layer 16 of OPT-6.7B, for two contexts at the same decoding step.}
\label{figs:keepnum-per-head}
\end{figure}

In the baseline full-attention mechanism, the sum of attention weights across all attended key entries within each attention head equals 1. Consequently, the cumulative attention value measures the effectiveness of sparse attention. Figure~\ref{figs:keepnum-per-head} shows the percentage of KV entries needed to achieve a 0.99 aggregated attention score at each of the 32 attention heads in layer 16 of the OPT-6.7B model. The figure includes the pattern of KV attention scores for two texts from the WikiText dataset, emulating two independent inference requests from users. As shown in Figure~\ref{figs:keepnum-per-head}, there is a large disparity in the distribution of attention scores at different heads. For example, while 10\% of the skewed KV entries account for 99\% of the attention scores at head 12, it requires close to 80\% of the uniformly distributed KVs at head 30 to achieve the same cumulative score. Furthermore, there is substantial variation in the attention distribution in different texts. These two observations suggest that sparse attention should be applied at a finer granularity, at the per-head level. A uniform layer-wise top-$k$ selection policy can result in the exclusion of critical KV entries in some heads and the inclusion of redundant ones in others.



\begin{tcolorbox}[colback=gray!10, colframe=gray!100, sharp corners, boxrule=0.0pt]
 {\bf O-1}: Attention distributions become increasingly skewed in deeper layers and exhibit substantial variation across different heads within the same layer.
 \end{tcolorbox}

\begin{figure}[t]
\setlength{\abovecaptionskip}{0.1cm}
\setlength{\belowcaptionskip}{-0.05cm}             
\centering
\includegraphics[width=0.9 \columnwidth]{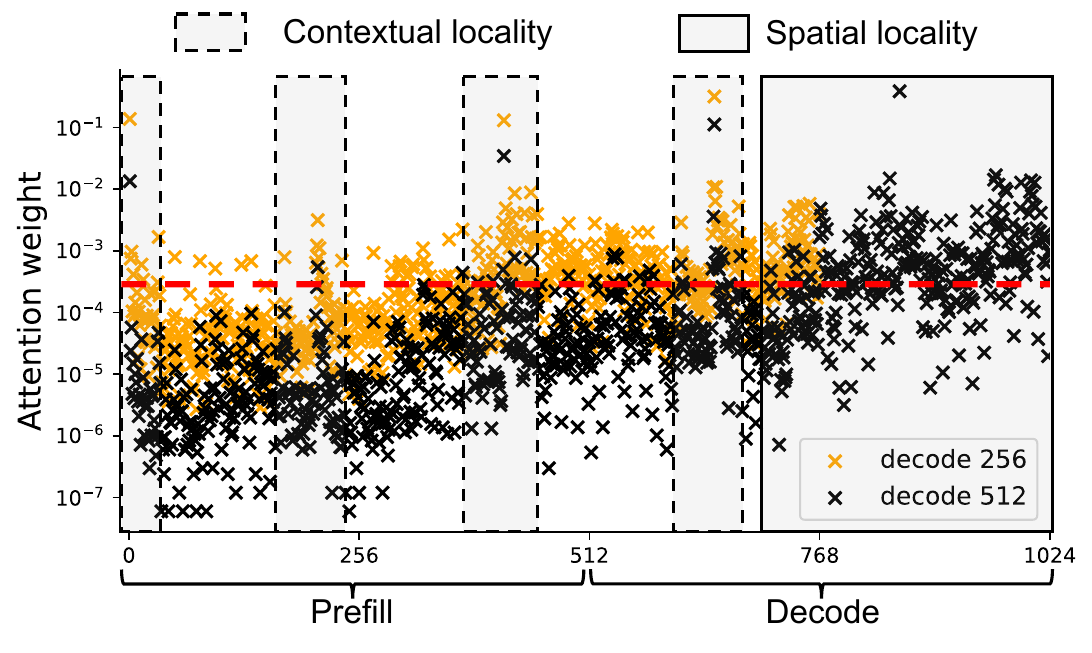}
\caption{Attention weights at Head 6 of Layer 16 with respect to cached key-value entries during decoding of the 256th and 512th tokens, given a prefill length of 512 tokens.}
\label{figs:attention-locality}
\end{figure}

Figure~\ref{figs:attention-locality} shows the distribution of attention scores for a single inference request at two decoding positions: the 256th and 512th tokens. The orange dots represent KV entries present after generating 256 tokens, while the black dots include additional entries that extend the KV cache to token 512. We make two observations. First, attention scores exhibit spatial locality, concentrating on recently computed KV entries (shown within the solid rectangle) as well as on the initial KVs from the prefill stage, both of which receive high attention scores. The scores also decay as tokens move away from the current token. Second, the attention score at the same position decreases as decoding progresses since the distance between the token and the decoding frontline increases. However, some critical KV entries exhibit strong contextual locality, remaining influential throughout the decoding process, as indicated by the few high-scoring orange and black dots within the dotted box. These are important tokens for this particular prompt. The red horizontal line delineates the set of KV entries required to reach a cumulative attention score of 0.9 for the latest output length. It can be clearly seen that the set of critical KVs contains most KVs within the spatial locality region and a few contextually important KVs generated at an earlier time.


\begin{tcolorbox}[colback=gray!10, colframe=gray!100, sharp corners, boxrule=0.0pt]
{\bf O-2}: Attention distributions exhibit spatial locality and contextual locality within a given context.\end{tcolorbox}  \label{obs:2}
 
\begin{figure}[!ptb]
\setlength{\abovecaptionskip}{0.1cm}
\setlength{\belowcaptionskip}{-0.05cm}             
\centering
\includegraphics[width=1.0 \columnwidth]{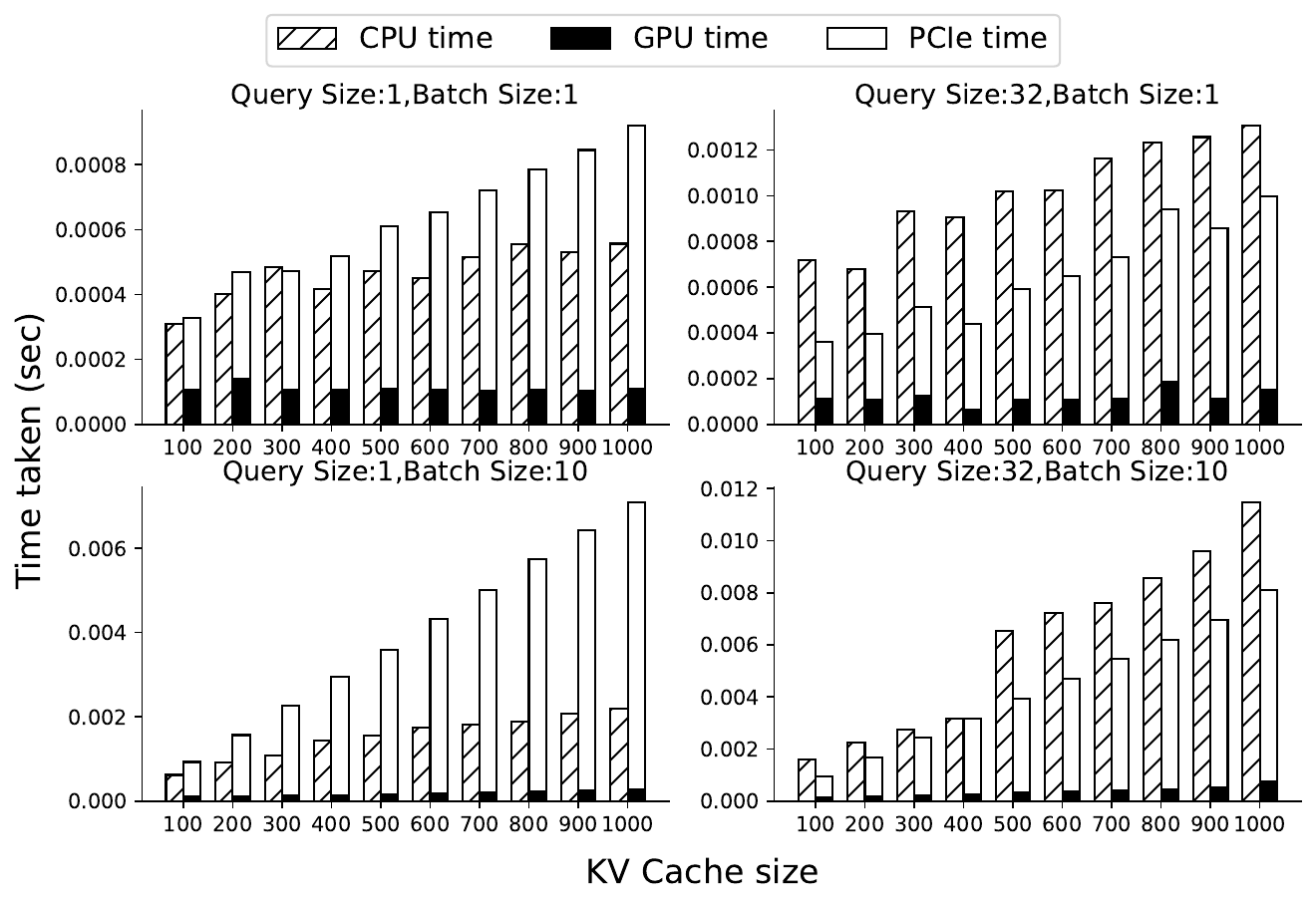}
\caption{ A breakdown of CPU-GPU attention time.}
\label{figs:attn-time-breakdown}
\end{figure}

Figure~\ref{figs:attn-time-breakdown} compares CPU-based and GPU-based attention mechanisms when {\em loading KV cache entries from host memory to the GPU}. Although GPUs generally offer higher compute throughput and memory bandwidth compared to CPUs, they do not always outperform CPUs in this setting. For single-token decoding (query size of 1), GPU-based attention is slower due to significant overhead from transferring KV cache entries between host and GPU memory. At a typical append operation size (query size of 32), GPU-based attention matches CPU performance as the computational overhead amortizes the transfer cost.  With larger batch sizes, GPU attention scales more efficiently due to its ability to exploit parallelism; however, the associated PCIe transfer overhead also increases proportionally. 
In all configurations, KV cache transfers over PCIe consistently dominate GPU attention latency, making them the primary performance bottleneck.


\begin{tcolorbox}[colback=gray!10, colframe=gray!100, sharp corners, boxrule=0.0pt]
 {\bf O-3}: CPU-based attention can achieve performance comparable to GPU-based attention when accounting for the overhead of loading KV entries into GPU memory over the PCIe bus.\end{tcolorbox} \label{obs:3}

\noindent {\bf Summary}. Our analysis of attention computation reveals a clear trade-off between CPU and GPU execution when the KV cache exceeds available GPU memory. We also demonstrate that fine-grained, per-head sparse attention has strong potential to deliver high accuracy. These observations motivated us to develop a hybrid CPU-GPU attention mechanism.


\subsection{More Related Work} 
\noindent{\bf Selective attention:} Selective attention to critical intermediate results is a widely studied strategy for mitigating the challenges of long-context inference in LLMs, especially under tight hardware constraints. It also improves model quality by discarding redundant or noisy KV cache entries. Prior work has shown that the importance of KV entries varies by task~\cite{beltagy2020longformer, kitaev2020reformer, zaheer2020big}, and has proposed efficient methods to retain only the most relevant ones. Owing to its success in NLP, selective attention has been extended to domains such as vision and graph processing~\cite{liu2021swin, yun2019graph, deng2024mega}. With the rise of chatbot applications and test-time scaling~\cite{muennighoff2025s1}, there is increasing demand for attention mechanisms that are both hardware-efficient and scalable. In response, recent work has introduced block-wise sparse attention kernels optimized for modern accelerators~\cite{yuan2025native, xu2025xattention, lu2025moba}.

\noindent{\bf KV cache management:} Efficient KV cache management is key to scalable LLM inference. Early approaches tackled GPU memory fragmentation via page-table-based cache management, as seen in \texttt{vllm}\cite{kwon2023efficient, prabhu2024vattention, adnan2024keyformer}, along with custom attention mechanisms. Later works proposed reusing KV entries across contexts to reduce redundancy\cite{yao2025cacheblend, gao2024cost}, and cache merging to lower memory usage~\cite{wang2024model, ge2023model}. KV reuse has been extended to distributed settings using disaggregated prefill-decode execution, where prefill-generated KV entries are transferred to decode clusters for improved utilization~\cite{agrawal2023sarathi, zhong2024distserve, qin2025mooncake}.

\noindent{\bf Hybrid computation:} The high compute and memory demands of long-context inference have spurred work on hybrid systems that leverage tiered memory and multiple compute devices~\cite{yu2024twinpilots, luo2025headinfer}. Some approaches dynamically scale models beyond device memory via fine-grained swapping~\cite{sheng2023flexgen} or computation partitioning~\cite{song2024powerinfer}. As longer contexts push KV cache sizes beyond on-device memory limits, emerging systems offload attention computations to external devices with larger memory~\cite{pan2024instinfer, ye2023accelerating, zhong2025hybrimoe}.

\input{design}
\input{impl}

\section{Evaluation}

\noindent\textbf{Platforms.}
All experiments were conducted on a dual-socket server, with each socket equipped with an Intel Xeon Gold 6430 processor (32 cores per socket). The system was configured with eight DDR5 memory modules, providing a total of 512 GB of system memory. The server also featured eight NVIDIA A6000 GPUs, each with 48 GB of device memory, interconnected via PCIe 4.0. Each GPU connection offers a maximum unidirectional bandwidth of up to 32 GB/s.

\noindent\textbf{Models and Workloads.}
We selected three representative open-source language models that vary in parameter scale and architectural design: LLaMA~\cite{touvron2023llama}, GPT-NeoX~\cite{black2022gpt}, and OPT~\cite{zhang2022opt}. To evaluate inference performance, we employed the WikiText dataset~\cite{merity2016pointer}, which is well-suited for assessing long-context generation capabilities.

\noindent\textbf{Baselines.}
We integrate \beyond into two widely used inference frameworks—FlexGen~\cite{sheng2023flexgen} and Hugging Face Transformers (HF)~\cite{wolf2019huggingface}, and evaluate its performance against relevant baselines. FlexGen targets memory-constrained, single-GPU environments and leverages a tiered memory hierarchy by swapping model components (e.g., weights and KV caches) between device and host memory. This design trades latency for extended capacity to maximize throughput. We compare \beyond to FlexGen-based alternatives that implement sparse attention mechanisms and KV offloading strategies, including Infinigen~\cite{lee2024infinigen}, which employs rehearsal-based KV prefetching across layers, and H2O~\cite{zhang2024h2o}, which evicts KV entries based on accumulated attention scores.

The Hugging Face Transformers (HF) framework offers broad support for diverse model configurations and features automatic device mapping when GPU memory is insufficient. We integrate \beyond with HF by managing KV cache placement in coordination with the framework’s device assignment logic. Performance is evaluated under two configurations: 1) a GPU-constrained setting where KV caches are offloaded, and 2) a GPU-rich setting where all KV caches are retained in high-bandwidth device memory.

\noindent\textbf{Performance Metrics.}
We report the following performance metrics: 1) end-to-end task completion time as a function of output sequence length and batch size, 2) peak memory consumption attributable to KV attention in addition to static model weights, 3) inference throughput across varying generation configurations, and 4) per-token generation latency. To ensure that \beyond’s modified attention mechanism does not compromise model accuracy, we also report perplexity scores, a measure of per-token generation accuracy according to the ground truth.

\begin{figure}[!ptb]
\setlength{\belowcaptionskip}{-0.2cm}           
\centering
\includegraphics[width=0.9  \columnwidth]{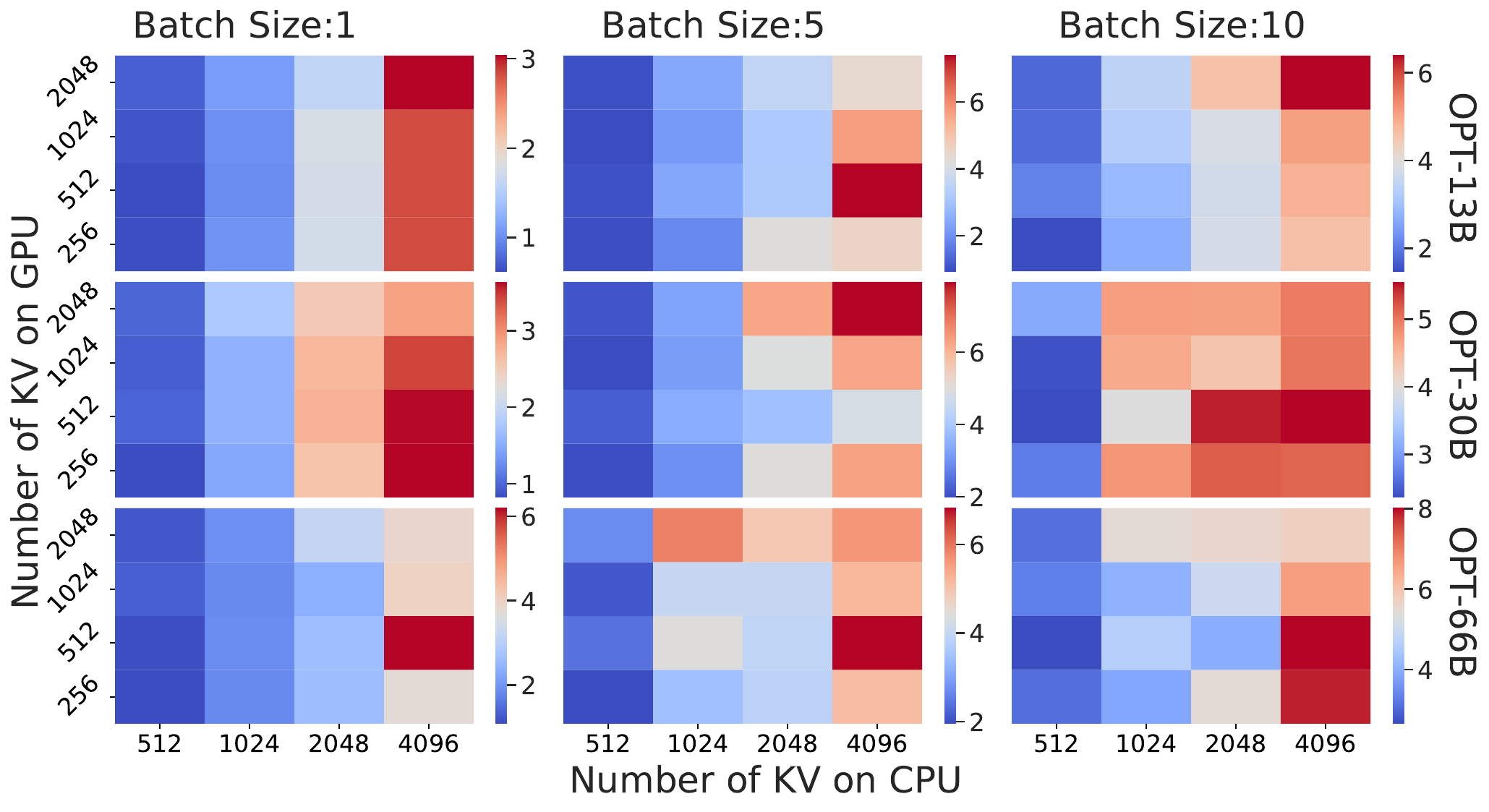}

\caption{Speedup due to \beyond's hybrid attention.}
\label{figs:hybrid-speedup-a6000}
\end{figure}

\subsection{Micro-benchmark} 
\begin{figure}[!ptb]
\setlength{\belowcaptionskip}{-0.2cm}           
\centering
\includegraphics[width=0.9  \columnwidth]{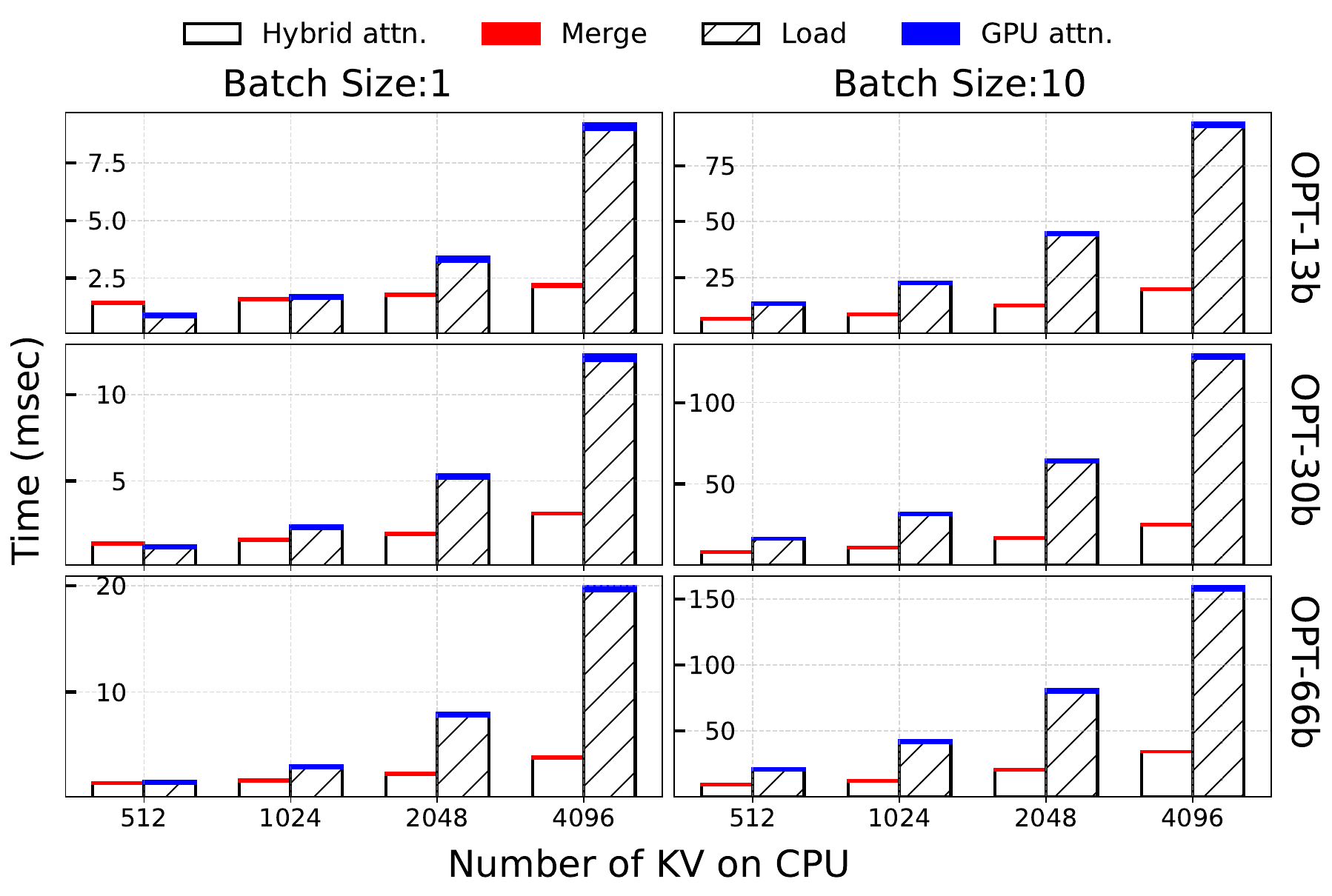}

\caption{A breakdown of attention time.}
\label{figs:breakdown}
\end{figure}

We begin by evaluating the benefits of \beyond’s hybrid attention on a single attention layer across varying batch sizes using three OPT LLMs. These models share the same hidden dimension size of 128 but differ in their number of attention heads. Figure~\ref{figs:hybrid-speedup-a6000} shows the speedup achieved by \beyond's hybrid attention compared to pure GPU attention, which requires loading KV entries currently resident in CPU memory into GPU memory. In comparison, hybrid attention computes attention on the CPU and only transfers the intermediate results to the GPU. The y-axis shows the number of KV entries already resident in GPU memory, while the x-axis shows those offloaded to CPU memory. A warmer color in the figure indicates a higher speedup due to hybrid attention. The figure indicates that hybrid attention becomes increasingly beneficial as more KV entries reside in CPU memory and batch size increases, making it particularly effective for long-context generation. 
Figure~\ref{figs:breakdown} further shows a breakdown of attention time in these two approaches, in which the number of KV entries on GPU was set to 1024. The figure suggests that PCIe transfer time for KV entries becomes an increasing bottleneck as KV size grows. Although CPU attention is several times slower than GPU attention, the {\em merge} time, which involves transferring CPU attention results over PCIe, is negligible, resulting in an overall speedup.


\subsection{End-to-End Model Performance}
\begin{figure}[!ptb]
\setlength{\belowcaptionskip}{-0.2cm}           
\centering
\includegraphics[width=1    \columnwidth]{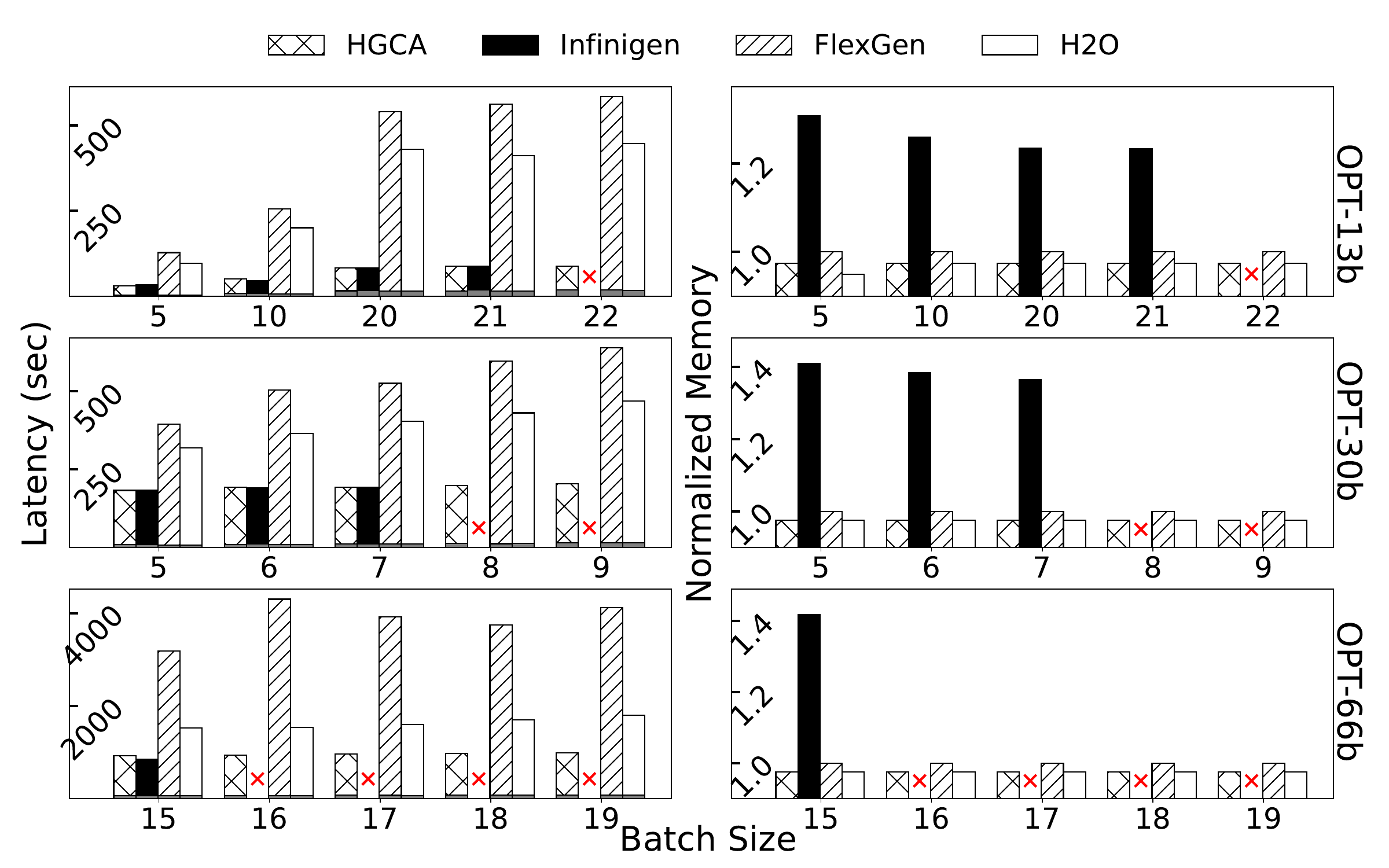}

\caption{Generation performance of OPT models using FlexGen under varying batch sizes and model configurations.}
\label{figs:flexgen_test}
\end{figure}

Next, we compare the end-to-end inference performance of \beyond against several representative approaches, focusing on throughput, task completion time, and memory usage. First, we leverage the {\bf FlexGen} framework to evaluate the inference performance on a single NVIDIA A6000 GPU, using the OPT model. We measured the total generation time of 128 tokens with a prefill length of 1920 tokens. FlexGen supports on-demand model weight swapping and hence can accommodate larger models. As shown in Figure~\ref{figs:flexgen_test}, we evaluated three OPT models and increased the batch size up to the maximum supported by each method -- FlexGen, H2O, Infinigen, and \beyond. Since the OPT-30B and OPT-66B models cannot fit into a single GPU, we statically configure 75\% of the model weights to reside on the GPU for OPT-30B, and 25\% for OPT-66B. Additionally, following the experimental setup in Infinigen~\cite{lee2024infinigen}, 100\% of the KV entries are initially placed in CPU memory and loaded into GPU memory on demand during attention. FlexGen performs full attention, whereas the other approaches use sparse attention. According to the settings in their original publication, we set Infinigen and H2O to perform sparse attention on 20\% of KVs (top-$k$). We set the percentage of KVs resident on GPU in \beyond to 5\%, with the remaining stored and computed on CPU.



Figure~\ref{figs:flexgen_test} shows that \beyond consistently outperformed FlexGen and H2O across all batch sizes, primarily due to the overhead in those approaches from KV loading and LRU-based KV eviction. In contrast, Infinigen achieved comparable performance but incurred significantly higher memory usage, as its attention rehearsals require additional memory buffers to predict attention scores. Unfortunately, due to its high memory overhead, Infinigen encountered out-of-memory (OOM) errors across all models, with failures particularly pronounced in the large OPT-66B model.

Hugging Face ({\bf HF}) provides an intuitive, user-friendly framework that supports automatic model-to-device mapping, compatible with a broader range of models. One important design in HF is automatic partitioning and allocation of model weights on multiple GPUs to assist model deployment and dynamic allocation of KV caches to utilize the remaining GPU memory. Unlike \beyond, HF performs full attention and can only scale by utilize multiple GPUs. In contrast, \beyond restricts the amount of GPU memory for KV storage and pre-allocates memory for on-GPU KV entries. We compared two variants of \beyond against HF. First, we scaled \beyond to use multiple GPUs following HF's partitioning of model weights but only enabled GPU (full) attention (denoted as GPU KV ratio: 1). Second, \beyond is configured to enable hybrid attention to fully utilize a single GPU before scaling to a second one (denoted as GPU KV ratio: 0.5).

\begin{figure}[!ptb]
\setlength{\belowcaptionskip}{-0.2cm}           
\centering
\includegraphics[width=0.95     \columnwidth]{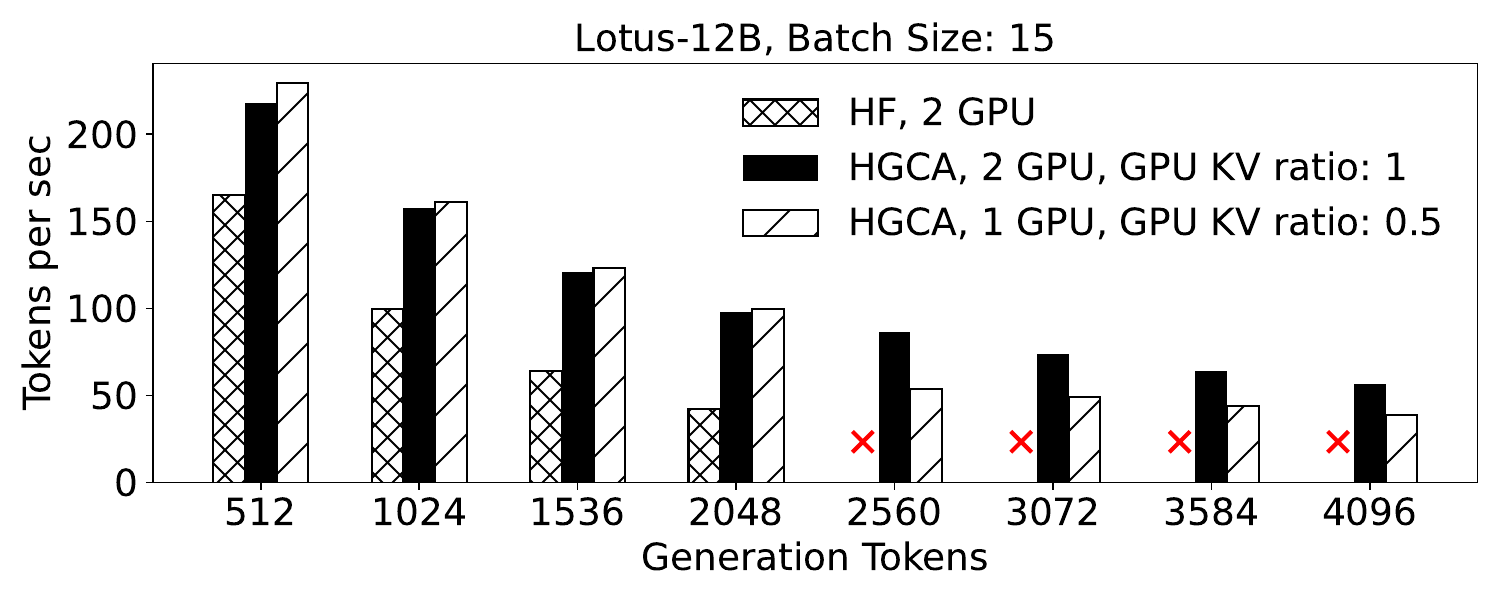}

\caption{The generation of 4096 tokens in GPT-Neox-12B.}
 
\label{figs:lotus_12b}
\end{figure}
\begin{figure}[!ptb]
\setlength{\belowcaptionskip}{-0.2cm}           
\centering
\includegraphics[width=0.95     \columnwidth]{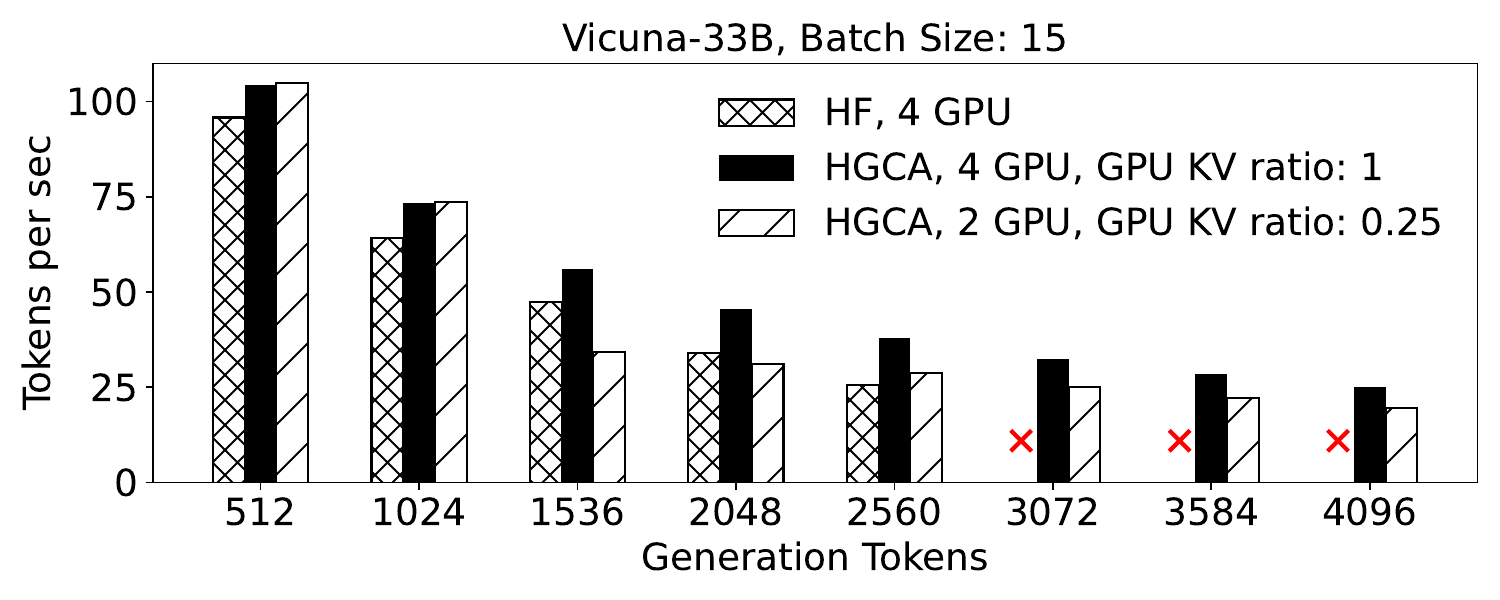}

\caption{The generation of 4096 tokens in Llama-33B.}
\label{figs:vicuna_33b}
\end{figure}

Figure~\ref{figs:lotus_12b} shows the performance of generating 4096 tokens using model GPT-Neox-12B. HF scaled this model to two A6000 GPUs and can accommodate the generation of up to 2048 tokens on these GPUs. In contrast, \beyond was capable of serving all tokens using a single GPU and hybrid attention. We made three observations. First, \beyond's GPU-only full attention outperformed HF's attention with dynamic KV allocation. \beyond's pre-allocation of KV caches avoided potential memory fragmentation and was more efficient. This highlights \beyond’s advantage in offloading KV entries without the risk of exhausting GPU memory so that users can comfortably reserve GPU memory for KV caches. Second, HF cannot scale beyond 2048 tokens due to the lack of KV offloading while \beyond was able to scale to the full generation length. Third and most importantly, while \beyond's hybrid attention results in a modest reduction in token throughput compared to its full GPU attention, it achieves this performance using only half the GPU resources. We further scaled the experiment to a larger Llama-33B model and 4 GPUs, as shown in Figure~\ref{figs:vicuna_33b}. The observations were similar except that \beyond's hybrid attention using 2 GPUs was less performant than its full attention using 4 GPUs. The performance discrepancy was due to the parallelization of inference involving model weights. With larger models, the gap between \beyond's full and hybrid attention narrowed towards the end of the long sequence generation.

\subsection{Accuracy}

\begin{table}[t]
\small
\centering
\setlength{\belowcaptionskip}{0pt}
\footnotesize
\caption{The comparison of perplexity between full attention and \beyond's hybrid attention. Text lengths are 2048, 4096, and 8192 for OPT, GPT-Neox, and LLaMA-based models, respectively. Reference perplexity and \beyond's better-than-reference performance are highlighted in bold.}
\label{tab:full_combined_ppl}
\vspace{1ex}
\begin{tabular}{c | c | c | c | c | c}
\toprule
\makecell{Model \\\textbf{Baseline PPL}} & \makecell{GPU \\ Ratio} & $\beta=0.25$ & $\beta=0.5$ & $\beta=0.75$ & $\beta=1.0$ \\
\midrule

\multirow{3}{*}{\makecell{OPT-6.7B\\\textbf{19.12}}} 
 & 0.25 & 19.42 & 19.17 & \textbf{19.02} & 19.12 \\
 & 0.5  & 19.17 & 19.12 & 19.09 & 19.09 \\
 & 0.75 & 19.12 & 19.17 & 19.17 & 19.20 \\
\midrule
\multirow{3}{*}{\makecell{OPT-13B\\\textbf{17.69}}} 
 & 0.25 & 17.97 & 18.05 & 18.08 & 17.94 \\
 & 0.5  & 17.77 & 17.80 & 17.80 & 17.83 \\
 & 0.75 & 17.69 & 17.69 & 17.69 & 17.72 \\
\midrule
\multirow{3}{*}{\makecell{OPT-30B\\\textbf{16.78}}} 
 & 0.25 & 19.97 & 19.92 & 19.78 & 19.70 \\
 & 0.5  & 16.91 & 16.78 & 16.78 & \textbf{16.75} \\
 & 0.75 & 18.33 & 27.83 & 19.42 & 18.47 \\
\midrule
\multirow{3}{*}{\makecell{LLaMA-2-7B\\\textbf{124.4}}} 
 & 0.25 & 112.1 & 115.9 & 106.3 & 105.4 \\
 & 0.5  & 105.3 & 100.8 & \textbf{93.28} & 99.81 \\
 & 0.75 & 124.6 & 124.9 & 125.6 & 125.5 \\
\midrule
\multirow{3}{*}{\makecell{LLaMA-2-13B\\\textbf{227.2}}} 
 & 0.25 & 288.7 & 308.9 & 310.2 & 307.2 \\
 & 0.5  & 293.8 & 308.8 & 330.9 & 341.1 \\
 & 0.75 & 226.2 & 225.7 & 225.6 & \textbf{225.1} \\
\midrule
\multirow{3}{*}{\makecell{Vicuna-33B-v1.3\\\textbf{889.1}}} 
 & 0.25 & 993.5 & 1013 & 941.5 & 983.5 \\
 & 0.5  & 896.2 & 985.4 & 894.9 & 893.8 \\
 & 0.75 & 1013 & 1033 & 1130 & 1013 \\
 \midrule
 \multirow{3}{*}{\makecell{GPT-NeoX-12B\\\textbf{66.88}}} 
 & 0.25 & 51.38 & 51.19 & 51.19 & \textbf{51.09} \\
 & 0.5  & 54.59 & 52.81 & 51.91 & 51.59 \\
 & 0.75 & 65.88 & 65.38 & 64.81 & 64.56 \\
 
\bottomrule
\end{tabular}
\end{table}

As one of the standard metrics in language modeling, perplexity quantifies how well a model predicts a token sequence, with lower values indicating better predictive performance. Evaluating perplexity over longer sequences provides deeper insight into the model’s ability to capture long-range dependencies. However, the existing work on inference serving lacks a comprehensive evaluation of perplexity on long sequence generation. For example, Infinigen only measured the perplexity of the first token generated rather than the full text. To this end, we compare \beyond's model accuracy with that of Hugging Face's full attention, which does not discard any KVs and is considered a reference performance.  

Recall that \beyond employs sparse attention on the CPU side that exploits contextual KV locality. We varied the filtering factor $\beta$ and the proportion of KV attention performed on the GPU, which is considered lossless. By adjusting $\beta$, we control how aggressively \beyond discards less important KV entries. A higher $\beta$ value indicates more selective filtering. As shown in Table~\ref{tab:full_combined_ppl}, \beyond's hybrid attention achieved near-identical model accuracy compared to full attention in all models, even outperforming full attention in certain cases. Most interestingly, the GPU KV ratio seemed to have no clear impact on accuracy, highlighting the effectiveness of sparse attention. Furthermore, \beyond's best model performance typically resulted in a large $\beta$ value and more selective KV filtering. This finding motivated us to pursue more aggressive sparse attention and further improve inference performance in resource-constrained environments, as a future work.  
\begin{figure}[!ptb]
\setlength{\belowcaptionskip}{-0.2cm}           
\centering
\includegraphics[width=1    \columnwidth]{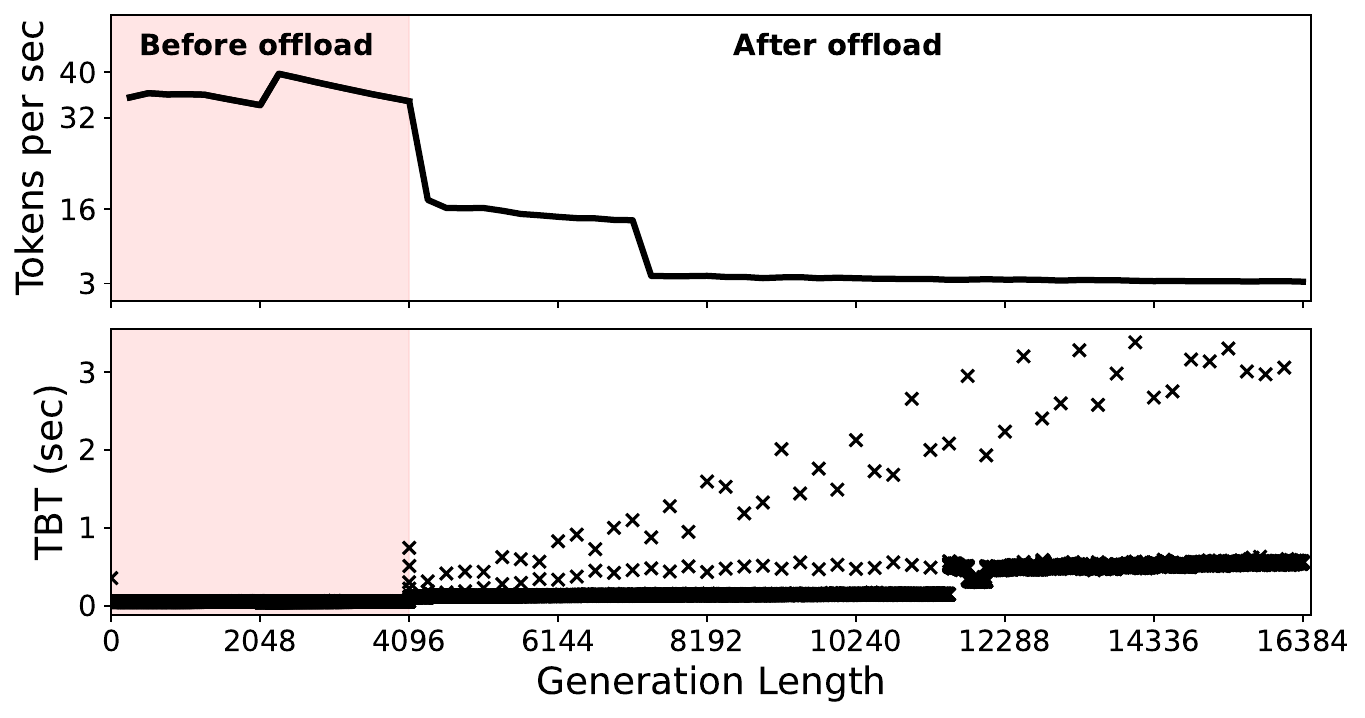}

\caption{Long context inference with GPT-Neox-12B.}
\label{figs:tbt_decode_16384}
\end{figure}




\subsection{Long Context Inference}
In this experiment, we measured \beyond's performance during decoding while allowing the KV cache to grow proportionally with the sequence length. The model was configured to perform continuous decoding for a sequence of 16,384 tokens within a single request (batch size 1), as illustrated in Figure~\ref{figs:tbt_decode_16384}. \beyond's KV size on GPU was configured to 4096 KVs and $\beta$ was set to 1. Hybrid attention was invoked once sequence length exceeded 4096. Figure~\ref{figs:tbt_decode_16384} shows both the token rate per second and time between tokens (TBT). \beyond's hybrid attention helped scale to a long sequence output without OOM errors with acceptable performance. For example, toward the end of generation, \beyond can still deliver 3–4 tokens per second, equivalent to 180–250 tokens per minute, which approaches the average human reading speed. However, we also observed large variations and outliers in TBT as decoding progressed. The problem was due to inefficient multi-thread scheduling and load imbalance during CPU attention. By demonstrating the potential of CPU sparse attention, we anticipate that this will inspire further research into optimizing sparse attention on CPUs.

\section{Conclusions}
We present \beyond, a novel hybrid CPU-GPU attention mechanism that enables scalable LLM inference by combining GPU-based dense attention with CPU-optimized sparse attention. \beyond preserves model accuracy while significantly reducing GPU memory pressure. Through efficient fusion and fine-grained, per-head sparsification, \beyond delivers strong performance gains and achieves superior scalability, enabling support for longer contexts and larger batch sizes on commodity GPU hardware.



\bibliographystyle{ACM-Reference-Format}
\bibliography{reference}




\end{document}

%% file: design.tex
\section{\beyond}
In this section, we introduce \beyond, a novel GPU-CPU hybrid attention mechanism designed to deliver high accuracy and efficiency in LLM inference, especially for long-sequence contexts and large batch sizes. We begin with a high-level system overview of \beyond and outline its core challenges and our solutions (\S\ref{sec:overview}). We then detail the key components of \beyond, including locality-aware KV cache management (\S\ref{sec:kv-manager}) along with a hybrid attention mechanism (\S\ref{sec:hybrid_attention}).

\begin{figure}[t]
\setlength{\belowcaptionskip}{-0.2cm}           
\centering
\includegraphics[width=0.9 \columnwidth]{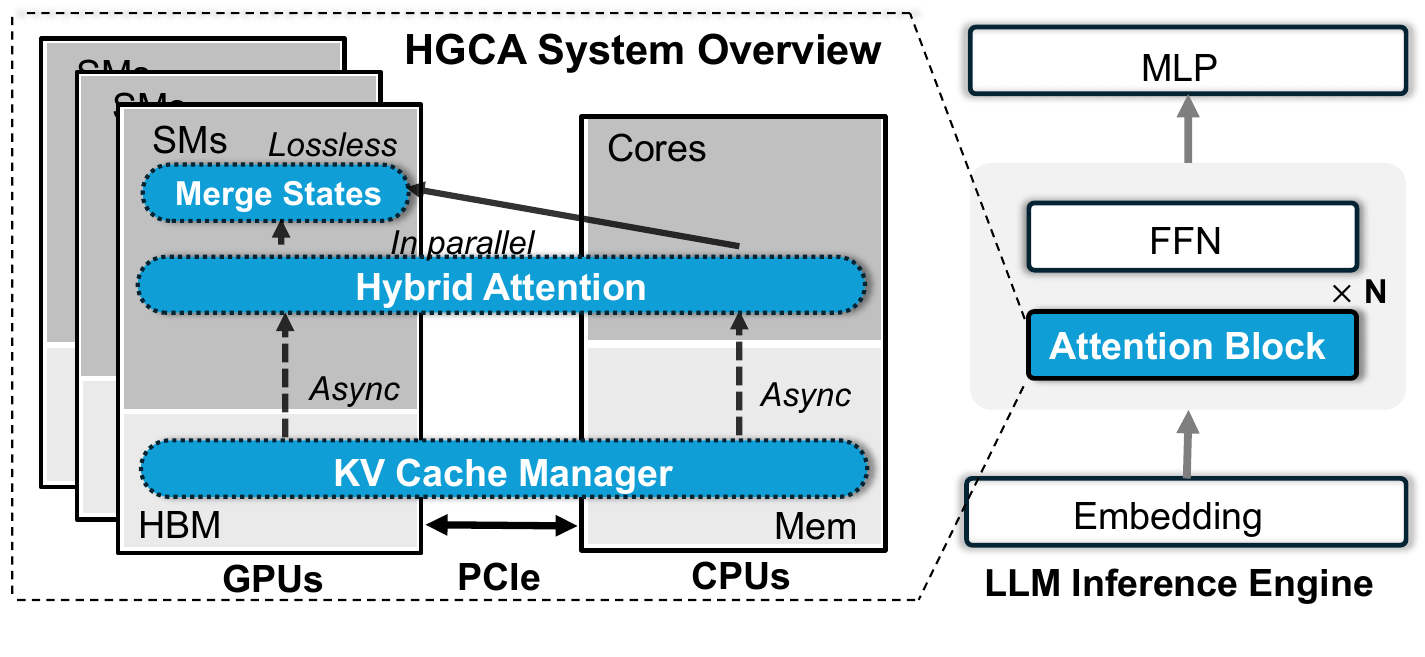}
\vspace{-3mm}

\caption{System overview of \beyond.}
\label{figs:beyond-overview}
\end{figure}
\subsection{System Overview}
\label{sec:overview}
Figure~\ref{figs:beyond-overview} illustrates the architecture of \beyond, which operates within the attention blocks of LLM inference. \beyond orchestrates memory and computation across heterogeneous hardware to address the growing memory footprint of KV caches in LLM serving. It selectively offloads portions of the KV cache to CPU memory, thereby overcoming GPU memory limitations for long sequences. Unlike prior offloading-only approaches (e.g., InfiniGen~\cite{lee2024infinigen} which extend memory capacity via CPU), \beyond goes further by harnessing CPU compute to cooperatively execute attention operations alongside the GPU. This co-design transforms a standard GPU-only inference pipeline into a collaborative CPU–GPU execution, enabling efficient and scalable LLM inference without sacrificing performance or accuracy.

However, achieving efficient joint attention execution on CPU and GPU requires overcoming two key challenges: (1) {\em Compute disparity}: Although equipped with high-capacity memory with comparable memory bandwidth to GPUs, CPUs provide significantly lower throughput for matrix-intensive workloads than modern GPUs. Naively offloading a large share of attention computation to the CPU would bottleneck the end-to-end throughput. The system must carefully balance the workload so that CPU computation on offloaded KV data does not slow down the GPU’s progress. (2) {\em Limited and slow interconnect}: The PCIe bus linking CPU and GPU has constrained bandwidth and added latency. Frequent or large-volume transfers of KV cache data can easily become a bottleneck, overwhelming the bus and stalling the GPU. Minimizing communication overhead is crucial to prevent severe throughput degradation in an offloading scenario. 

To address these, \beyond introduces two key techniques -- i.e., {\em locality-aware KV cache management} (\S\ref{sec:kv-manager}) and {\em bandwidth-efficient hybrid attention} (\S\ref{sec:hybrid_attention}) -- jointly manage where data lives and how attention is computed.

First, \beyond's KV cache manager allocates and manages the KV cache across the GPU and CPU, guided by the observations that recently generated tokens have the highest attention relevance for the next token (\S\ref{sec:motive}). The manager prioritizes keeping recent KV entries in GPU memory and opportunistically offloads older, less influential entries to CPU memory when GPU memory becomes constrained. This recency-aware placement means that at each decode step, the GPU has fast access to the most pertinent context, while the bulk of infrequently used context is stored in ample CPU memory. More importantly, this recency-aware offloading enables \beyond to better exploit the {\em sparsity} in attention calculation for distant tokens. Since those tokens usually contribute low weights to final attention, their KV entries can be processed efficiently with sparse attention on CPUs. Using CPUs' advanced control logic and multicore architectures, \beyond invents a finer-grained, per-head sparsification approach, significantly reducing computational load without impacting accuracy. This sparsification substantially narrows the performance gap between CPUs and GPUs. 

Further, inspired by the tiled attention computation~\cite{dao2022flashattention}, \beyond executes attention in two coordinated parts and then combines the results with a {\em lossless} aggregation. It treats the GPU-resident and CPU-resident KV cache entries as two separate blocks for attention calculation. The GPU computes attention over its local block of recent KV entries (with full attention), while the CPU in parallel computes attention over the offloaded block of older entries (with sparse approximations). Each side produces a partial result along with the necessary normalization statistics, with which \beyond applies normalization to merge these partial attention outputs on the GPU. The key insight is that the intermediate outputs needed to merge results are extremely small -- essentially the partial attention outputs and normalization statistics transferred from the CPU to the GPU over PCIe, i.e., orders of magnitude smaller than transferring the entire offloaded KV cache. The final attention context reflects contributions from the full sequence, i.e., both recent and distant tokens.

In summary, \beyond uses CPUs as a large-memory computational partner rather than a mere storage device, overlapping its work with GPU computation. By integrating locality-aware KV placement with a bandwidth-efficient hybrid attention mechanism, \beyond alleviates GPU memory pressure, minimizes PCIe traffic, and enables scalable, high-throughput inference for long-sequence contexts and large batch sizes with a single consumer-class GPU.

\begin{figure}[t]
\setlength{\belowcaptionskip}{-0.2cm}           
\centering
\includegraphics[width=1 \columnwidth]{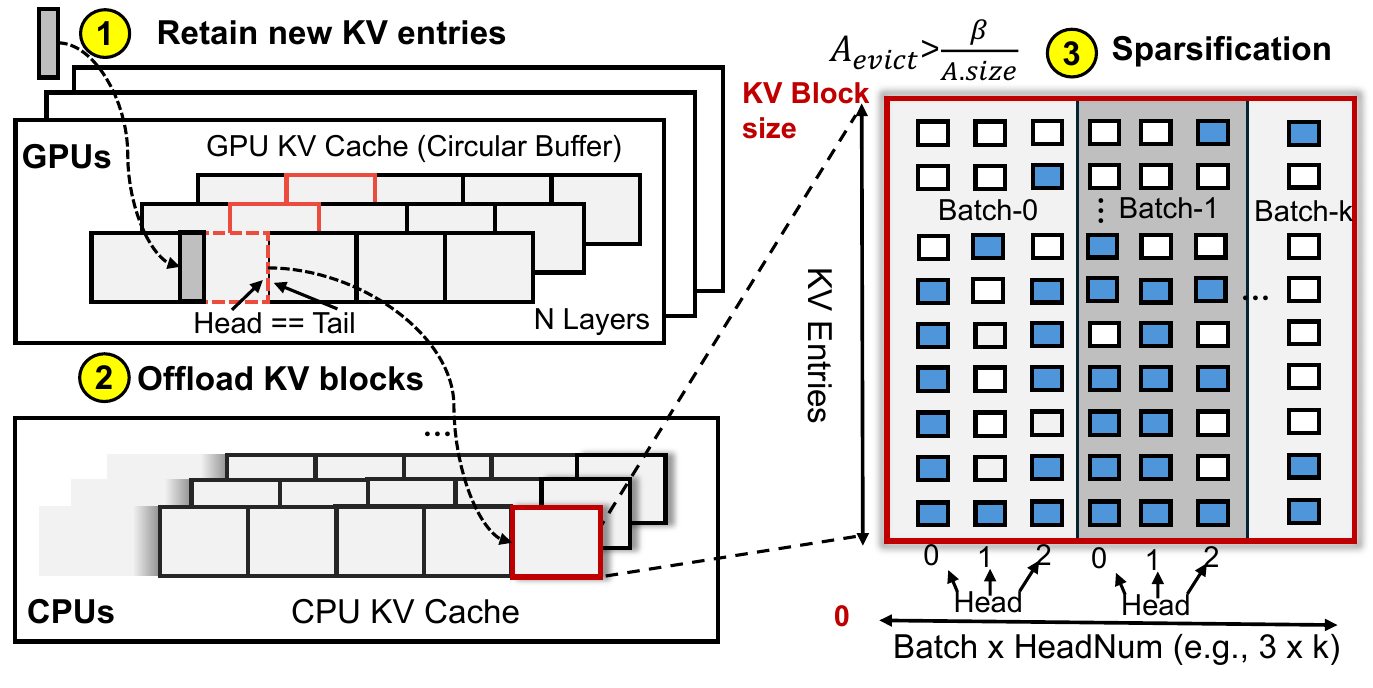}

\caption{\beyond: KV cache manager.}
\label{figs:kv-manager}
\vspace{-8pt}
\end{figure}

\begin{algorithm} [t]
\caption{KV manager.}\label{alg:kv-manager}
\begin{algorithmic}[1]
\footnotesize
\Require  $KV _{in}, A _{gpu},A_{cpu}[optional] $

\State \textit{\color{orange}//GPU side management w/ async stream}
\State Defined: \textit{ $blk\_num$, $blk\_size$ , $\alpha$, $\beta$    } \label{line:window-coverage}
 
\State $KV^i_{gpu} \gets$  \text{GPU KV cache at $i^{th}$ layer} 
 
\State $l_{max} \gets$ $blk\_num$ * $blk\_size$
\State $l_{cur} \gets$ $KV^i_{gpu}.size +KV_{in}.size$

\State $\alpha  \gets$ \text{Moving average factor} 
\State $A^i_{maw}  \gets$  \text{Moving average of attention weight at $i^{th}$ layer} 
\State$A^i_{maw} \gets$  $(1-\alpha) \times A^i_{maw} + \alpha \times A_{gpu}$ \label{line:maw}
\State $KV^i_{gpu}.add(KV_{in})$  

\If{$l_{cur} >= l_{max}$}
      
    \State  $l_{evict} =  \lceil  \frac{l_{max}- l_{cur}+1}{blk\_size}\rceil \times {blk\_size}$
    \State  $KV _{evict},A _{evict} \gets  KV^i_{gpu}[:l_{evict}],A^i_{maw}[:l_{evict}] $
     \State  $KV_{evict},A_{evict} \gets KV_{evict}.to(cpu),A_{evict}.to(cpu)$
     
\EndIf

\State \textit{\color{orange}//CPU side management w/ async thread}
\State $KV^i_{cpu} \gets$  \text{CPU KV cache at $i^{th}$ layer} 
\State $KV^i_{ctx} \gets$  \text{Contextual kv cache  at $i^{th}$ layer} 
 
\State $\beta  \gets$ \text{Sparsification factor}   
\If{ ($append$)  and (  $A^i_{cpu}$ is not None)} \label{line:re-evaluation}
    \State \Tab $ids  \gets (A^i_{cpu}>\frac{\beta}{=A^i_{cpu}.size}) $
    \State \Tab $KV^i_{ctx}  \gets KV^i_{cpu}[ids]$
\EndIf
\State  $ids  \gets (A_{evict}>\frac{\beta}{=A^i_{gpu}.size}) $
\State  $KV^i_{ctx}.add(KV _{evict}[ids])$
\State  $KV^i_{cpu}.add(KV _{evict})$

\State \Return  
\end{algorithmic}
\end{algorithm}

  

\subsection{Locality-Aware KV Cache Management}
\label{sec:kv-manager}
\beyond's KV cache manager focuses on two main tasks: (1) managing GPU and CPU memory for KV cache retention and offloading (\blacknumber{1} and \blacknumber{2}in Figure~\ref{figs:kv-manager}) and (2) tracking the relevance of KV cache for CPU-side sparsification (\blacknumber{3}).

\subsubsection{GPU-managed KV retention \& offload} 
\label{sec:gpukv}
The KV cache manager in \beyond pre-allocates a contiguous memory buffer on each GPU device to serve as a local KV cache pool, exclusively holding the KV entries generated by that GPU’s attention layers. This avoids frequent GPU memory allocations and fragmentation by providing a dedicated slab of memory per device. To align the cache layout with transformer-based LLM architecture -- i.e., composed of interleaved attention and feed-forward layers -- each GPU’s buffer is further partitioned into sub-regions corresponding to the model’s attention layers, i.e., for a transformer with $N$ attention layers, the GPU’s KV pool is divided into $N$ fixed-size segments, each caching one layer’s attention keys and values. This layer-wise partitioning mirrors the structure of the model and allows independent management of each layer’s KV entries (e.g., eviction and offloading can occur on a per-layer basis without affecting others). 

To exploit the {\em temporal locality} inherent in decoder attention (\S~\ref{sec:motive}, O-3), each layer-specific sub-region is organized as a {\em circular buffer} composed of fixed-size KV blocks~\footnote{\beyond batches KV offloads at the block granularity rather than per token, thus amortizing the cost of data movement and maximizing interconnect bandwidth utilization (e.g., PCIe). This batched eviction approach trades minor memory slack for significantly lower communication overhead, yielding higher throughput and more efficient CPU–GPU data transfers.}. Upon generation of a new token, its KV entry is inserted at the head of the buffer (Algorithm~\ref{alg:kv-manager}, \emph{line 9}), based on the insight that recently generated tokens are more likely to be attended in near-future decoding steps. As new entries accumulate, the buffer naturally advances in a FIFO manner, gradually overwriting older blocks. When the buffer reaches capacity, the oldest block located at the tail is selected for eviction and offloaded to CPU memory (\emph{lines 13}). This maintains the most relevant context within GPU memory for full attention, while less referenced or stale context is offloaded to CPU memory, where sparse attention mechanisms can be applied.

Finally, to enable fine-grained {\em relevance estimation} of the offloaded context, \beyond associates each KV entry with a moving average of attention weights (MAW) reflecting its historical importance during decoding. Specifically, each GPU-resident KV entry tracks a MAW value, updated throughout the generation process (\emph{line 8}), capturing how strongly it has been attended to. Upon eviction, the corresponding MAW is transferred alongside the KV block to CPU memory (\emph{line 13}). This lightweight metadata incurs minimal overhead on the GPU, yet allows the CPU-side runtime to perform attention-aware sparsification, selectively retaining or pruning context based on empirical utility rather than age alone.

\subsubsection{CPU-managed KV sparsification} 
\label{sec:cpusparse}
In contrast to GPU-local buffers, the host-side cache manager maintains references to the KV caches across all attention layers executed on the GPUs. It is architected to support dynamic expansion, thereby accommodating the storage of offloaded KV blocks (Figure~\ref{figs:kv-manager}). From these offloaded blocks, \beyond dynamically identifies and retrieves the most salient KV entries to facilitate efficient sparse attention computation.

Existing GPU-side sparsification approaches often rely on a fixed or predetermined attention budget to filter KV entries, e.g., H2O~\cite{zhang2024h2o} retains the top ~20\% of tokens with the highest cumulative attention scores. While such top-k token retention schemes are effective in reducing memory usage and speeding up attention computation, they are limited by GPU hardware constraints: GPUs excel at uniform, regular workloads with coalesced memory access and simple control flow; conversely, irregular, fine-grained operations are inefficient on GPUs, making more adaptive GPU-side methods that require additional runtime analysis incur high overhead due to added profiling and complex control logic~\cite{ge2023model, li2024snapkv}. 
Unfortunately, coarse-grained fixed-budget strategies can overlook subtle but important attention patterns (\S\ref{sec:motive}).

\noindent{\bf Head-granularity sparsification:} In contrast, \beyond offloads the KV cache to CPU memory and performs sparsification on the CPU side, where irregular control flow and fine-grained filtering can be handled efficiently. 

Immediately after offloading KV blocks and their MAW values to the host, an asynchronous CPU-side {\em sparsification} thread begins token selection {\em in parallel} with GPU inference. The CPU’s flexibility allows token selection at {\em per-head granularity} (motivated by O-2, \S\ref{sec:motive}), a fine-grained level difficult to achieve on GPUs. 
In Figure~\ref{figs:kv-manager} (\blacknumber{3}), \beyond evaluates each attention head’s KV entries independently, applying an adaptive thresholding mechanism to decide which entries to keep.
Let each KV entry’s accumulated attention weight be $A_{evict}$ (its MAW value) and let $A.size$ denote the total number of GPU-side KV entries.
We define a tunable threshold parameter $\beta$. A KV entry is selected for sparse attention if and only if it satisfies: $ A_{evict} > \frac{\beta}{A_{gpu}.size}$, i.e., its attention weight exceeds a fraction $\beta$ of the total attention count for that head. Entries meeting this criterion are deemed ``salient'' and kept, while those below the threshold are dropped from sparse attention. But they are still kept in the CPU memory buffer for future re-evaluation. After selection, the retained entries’ MAW values are re-normalized so that their attention weights sum to 1 for that head, preserving a valid probability distribution for downstream sparse attention computations.

The threshold parameter $\beta$ controls the aggressiveness of sparsification: a larger $\beta$ enforces stricter filtering, reducing the number of retained KV entries to lower CPU computation overhead -- potentially at the cost of accuracy -- while a smaller $\beta$ retains more entries to favor accuracy at the expense of higher overhead. This adaptive threshold operates per attention head, dynamically reflecting the head’s attention distribution: heads with sharp, peaked attention retain fewer high-value KV entries, whereas heads with flatter distributions preserve more. 

Empirical evaluation with $\beta$ typically set to 1 shows that \beyond’s per-head sparsification can select up to 30\% of tokens for some heads, and as few as less than 1\% for others -- while maintaining accuracy comparable to full attention. By evaluating each token’s MAW in the context of its corresponding head, \beyond’s CPU-based strategy captures context locality, preserving KV entries with prominent attention spikes, as illustrated in Figure~\ref{figs:attention-locality}.

\noindent\textbf{Re-evaluation:} Once KV entries are sparsified on the CPU side, they typically do not require {\em re-evaluation} until the current decoding session concludes, as the contextual locality has already been established. However, when a new prompt is issued -- commonly referred to as an append operation (e.g., in multi-turn interactions) -- the newly appended tokens may alter the contextual relevance of previously offloaded entries. Tokens that were previously considered insignificant may become salient as the context evolves, while previously important tokens may diminish in relevance. For example, Figure~\ref{figs:keepnum-per-head} demonstrates that the set of salient KV entries can vary significantly across different contextual conditions.

To accommodate such changes, an asynchronous evaluation thread is launched on the CPU following each append operation. This thread re-evaluates the contextual relevance of all KV entries stored on the CPU, as described in Line~\ref{line:re-evaluation}, using the attention scores $A_{cpu}$ computed from the complete CPU-side KV cache (Algorithm~\ref{alg:hybrid-attention}, line 4). KV entries that no longer meet the salience criterion—defined as $A_{evict} > \frac{\beta}{A_{cpu}.size}$—are evicted, while previously pruned entries that now exceed the threshold are reinstated. This dynamic re-sparsification mechanism ensures that the CPU-side KV cache remains adaptive to the evolving context, thereby preserving computational efficiency and enabling the model to incorporate newly relevant information.

\begin{figure}[ptb!]
\centering
\includegraphics[width=\linewidth]{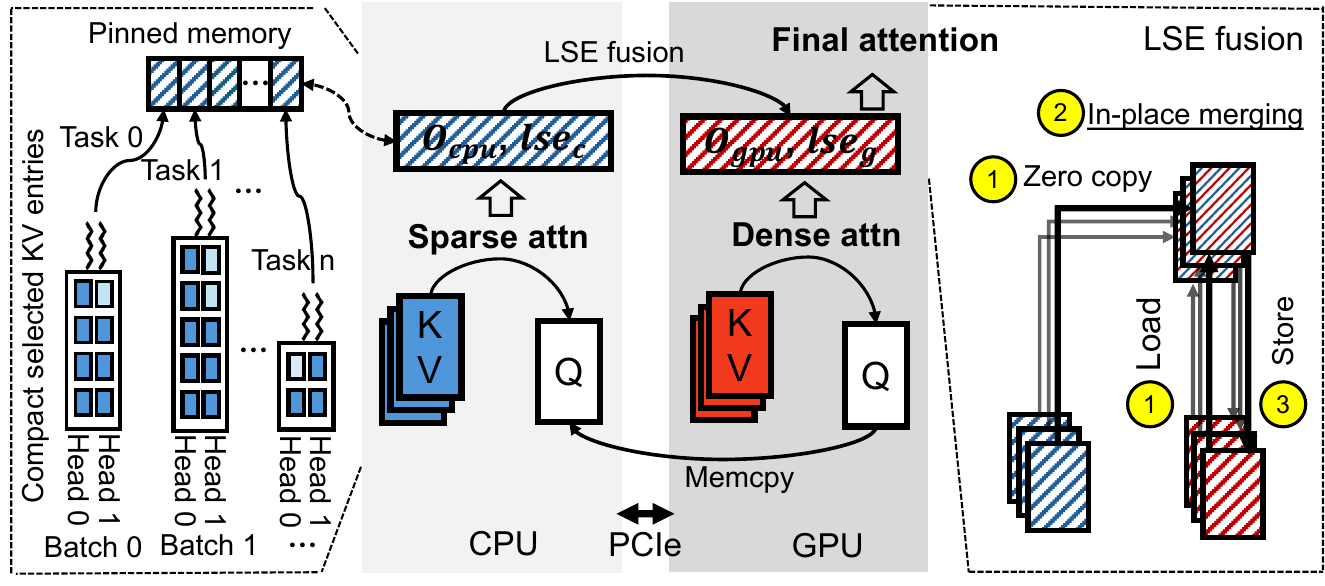}

\caption {\beyond: hybrid attention computation.}
\label{figs:workflow}
\end{figure}
\subsection{Hybrid Attention}
\label{sec:hybrid_attention}
\beyond introduces a joint hybrid attention mechanism, as shown in Figure~\ref{figs:workflow}, that splits the attention computation between the GPU and CPU. The GPU handles a sliding window of recent context with full, dense attention, while the CPU manages offloaded older context with sparse, selective attention. The final attention is obtained by merging these two partial attention results from the CPU and GPU. This co-design pursues two key goals: (1) {\em Reducing compute disparity}: The GPU performs heavy-density computation on a small subset, and the CPU handles only a lightweight sparse subset, preventing one from becoming a bottleneck. (2) {\em Enabling efficient collaboration:} GPU and CPU workloads execute concurrently, employing multicore parallelism (e.g., mapping sparse attention tasks across CPU cores), efficient data representations (e.g., local attention outputs and scaling factors for result merging), and communication optimizations (e.g., zero-copy data transfers). With these, \beyond maximally exploits the computational capabilities of CPU processors and effectively masks transfer latency.

\noindent{\bf GPU-local dense attention.} 
\beyond's KV cache management (\S~\ref{sec:kv-manager}) enables sliding-window dense attention on the GPU: it maintains only the most recent KV blocks in GPU memory, evicting older blocks to CPU memory as new tokens are generated. This design forms a bounded window over the latest $W$ tokens, which are processed using standard dense attention operations (Algorithm~\ref{alg:hybrid-attention}, line~\ref{line:gpu-attention}).

\begin{algorithm}
\footnotesize
\caption{Hybrid Attention}\label{alg:hybrid-attention}
\begin{algorithmic}[1]
\Require  $Q,KV_{in},layer\_idx $
\If{$decode$} \label{line:re-evaluation-load} 
     \State \Tab $KV_{gpu},KV_{ctx} = \text{kv\_manager.load($layer\_idx$)} $ 
\ElsIf{$append$}  
    \State \Tab $KV_{gpu},KV_{cpu} = \text{kv\_manager.load($layer\_idx$)} $ 
\EndIf

\State Launch async CPU tasks
\State \Tab  $task^i$ = cpu\_attention($Q[i],KV_{cpu||ctx}[i] $)  \label{line:cpu-attention}
\State Within GPU Stream
\State \Tab$KV_{gpu} \gets$ concat$(KV_{gpu},KV_{in})$
\State \Tab$O_{gpu},lse_g,A_{gpu}$ = gpu\_attention($Q,KV_{gpu}$) \label{line:gpu-attention}
\State Sync CPU tasks
\State \Tab$O_{cpu}[i],lse_c[i],A_{cpu}[i]  = task^i$ \label{line:cpu-attention}
\State $O =$ merge\_state($O_{cpu},lse_{c},O_{gpu},lse_{g}$)  \label{line:merge-results}
\State \Return  $O,A_{gpu},A_{cpu}$
\end{algorithmic}
\end{algorithm}

\noindent{\bf CPU-local sparse attention.} Recall that \beyond employs a per-head sparsification mechanism to select the most salient KV entries for sparse attention (\S~\ref{sec:cpusparse}). To enable efficient sparse attention computation, \beyond organizes these selected KV entries by attention head, storing each head's entries contiguously in CPU memory~\footnote{This reorganization is performed during sparsification and thus does not affect the critical path of attention computation.}. This contiguous arrangement results in compact subsets of KV entries per head, facilitating efficient parallel execution. Further, \beyond exploits the multiple cores of modern CPUs by distributing these subsets across multiple threads, e.g., each running on a separate CPU core. Each thread computes the attention exclusively for its assigned subset corresponding to a particular attention head. Since the subsets processed by different threads are disjoint, the final attention output is obtained by concatenating each thread's computed results, resulting in the combined multi-head attention output (e.g., stored in pinned memory as depicted in Figure~\ref{figs:workflow}).

However, increasing the batch size introduces a practical challenge: the number of required threads grows proportionally to the product of batch size and the number of attention heads, potentially leading to CPU oversubscription -- i.e., having much more threads than available CPU cores (i.e., increased context-switch overhead). To mitigate this issue, \beyond merges computations from multiple adjacent heads into a single sparse attention task, thereby reducing the overall thread count (Figure~\ref{figs:workflow})~\footnote{Practically, we choose the number of heads combined per task to be close to $ \frac{batch\_size \times head\_num}{core\_cnt}$.}. Combining multiple adjacent heads necessitates a relaxed per-head sparsification approach, as different heads may select varying numbers of KV entries. To align the selected KV entries across merged heads, \beyond permits padding with additional KV entries, even if they do not satisfy the threshold condition. CPUs' flexible and efficient control logic enables \beyond to easily manage this merging process and handle tasks of varying sizes -- a task substantially more challenging on GPUs, which favor rigid, uniform execution patterns and typically perform best on tasks of equal size.

 \noindent\textbf{Merging states.} After computing attention over the GPU sliding window and the selected CPU KV entries, \beyond merges the two partial results to obtain the final attention output. The objective is to combine these computations as if they were derived from a single softmax over the union of tokens, while preserving numerical stability and computational efficiency.
Inspired by FlashAttention~\cite{dao2022flashattention}, \beyond employs an {\em in-place aggregation and log-sum-exp (LSE) fusion} strategy. After the GPU computes its partial attention output $O_{\text{gpu}}$ 
(Algorithm~\ref{alg:hybrid-attention}, \emph{line 10}), and the CPU computes its partial output $O_{\text{cpu}}$ 
(\emph{line 12}), both outputs are locally \emph{normalized} within their respective domains and must be appropriately re-scaled to obtain a unified result.
Each partial result is associated with a log-sum-exp term of the form:
\[
\text{lse}_I = \log\left(\sum_{j \in I} e^{s_j}\right),
\]
where  $s_j = q \cdot k_j$ denotes the attention score corresponding to the $j^{th}$ key. The index set $I$ identifies the subset of tokens residing in a particular compute domain (CPU or GPU). 
A coordinated normalization step is then carried out to compute a unified scaling factor that spans both domains:
\[
z = e^{\text{lse}_\text{c}} + e^{\text{lse}_\text{g}}.
\]
Using this shared factor $z$ -- which is derived by merging two scalars between the CPU and GPU (the maximum score and the sum of exponentials) -- the partial outputs are scaled and combined in-place:
\[
\mathbf{O} = \frac{1}{z} \left(e^{\text{lse}_\text{c}} \cdot O_{\text{cpu}} + e^{\text{lse}_\text{g}} \cdot O_{\text{gpu}}\right).
\]
This yields an output equivalent to that produced by a single softmax operation over the union of GPU and CPU tokens. The merge is performed in-place by accumulating $O_{\text{cpu}}$ directly into the GPU’s output buffer and applying the shared normalization factor. The use of the LSE trick guarantees numerical stability when fusing attention scores from heterogeneous sources.
The communication overhead is minimal, as only the vector $O_{\text{cpu}}$ and scalar $\text{lse}_\text{cpu}$ are transmitted from the CPU to the GPU, avoiding the need to transfer full KV tensors. In practice, \beyond facilitates this exchange via {\em zero-copy} memory access, enabling the GPU to directly access CPU-resident data. This zero-copy strategy bypasses the GPU memory hierarchy and is especially effective for small or irregular data structures, making it well-suited for the dynamic behavior of \beyond's hybrid attention mechanism. The merging process occurs at the end of the attention layer, ensuring that the combined output can be directly consumed by the subsequent feed-forward network.

%% file: impl.tex
\section{Implementation}
We implemented \beyond using approximately 1.5K lines of Python code, based on PyTorch v2.4+cu124. 
 
\noindent{\bf KV Cache Manager:} The KV cache manager handles cache allocation and movement across CPU and GPU.
{\em On the GPU}, \beyond statically maps attention layers to devices at initialization and pre-allocates a contiguous buffer per device via \texttt{torch.empty}. Each buffer, managed by a dedicated CUDA stream, stores KV entries and their MAW metadata. \textit{Layer-level pointers} track start/end offsets for each batch, enabling efficient prefix access, incremental updates, and eviction detection. When capacity is exceeded, the oldest blocks are offloaded asynchronously to CPU memory via the assigned stream, avoiding disruption to inference.
{\em On the CPU}, \beyond uses \texttt{threading.Thread} to manage per-layer KV caches as growable tensor lists. A \textit{context cache} retains selected KV entries per head in separate tensors, allowing variable-length retention. A dedicated thread performs sparsification to update this cache.
Prior to each query, CUDA streams are synchronized to finalize GPU updates. At each layer’s end, CPU threads synchronize to ensure filtered blocks are merged. All operations remain off the inference critical path.

\noindent{\bf Hybrid Attention:} \beyond's hybrid attention leverages \texttt{torch.jit} to optimize execution across GPU and CPU. On GPU, dense attention integrates with minimal modifications to compute log-sum-exp (LSE) statistics. On CPU, sparse attention over the contextual KV cache is parallelized via \texttt{torch.fork}, assigning each thread a subset of heads. Despite variable context lengths, fixed output shapes allow pre-allocation of a shared pinned buffer, with threads writing to precomputed offsets. \texttt{torch.wait} ensures synchronization before merging. We extend \texttt{merge\_state} from \textsc{FlashInfer}~\cite{ye2024cascade} for zero-copy CPU-GPU fusion, and support batched merging along batch and head dimensions for scalability.